# Deep Epitome for Unravelling Generalized Hamming Network: A Fuzzy Logic Interpretation of Deep Learning

**Lixin Fan**
Nokia Technologies
Tampere, 33100, Finland
`lixin.fan@nokia.com`

## Abstract

This paper gives a rigorous analysis of trained Generalized Hamming Networks (GHN) proposed by Fan (2017) and discloses an interesting finding about GHNs, i.e., *stacked convolution layers in a GHN is equivalent to a single yet wide convolution layer*. The revealed equivalence, on the theoretical side, can be regarded as a constructive manifestation of the *universal approximation theorem* Cybenko (1989); Hornik (1991). In practice, it has profound and multi-fold implications. For network visualization, the constructed *deep epitomes* at each layer provide a visualization of network internal representation that does not rely on the input data. Moreover, deep epitomes allows the direct extraction of features in just one step, without resorting to regularized optimizations used in existing visualization tools.

## 1 Introduction

Despite the great success in recent years, neural networks have long been criticized for their black-box natures and the lack of comprehensive understanding of underlying mechanisms e.g. in Bentez et al. (1997); Gülçehre & Bengio (2013); Shwartz-Ziv & Tishby (2017); Shrikumar et al. (2017). The earliest effort to interpret neural computing in terms of logic inferencing indeed dated back to the seminal paper of Mcculloch & Pitts (1943), followed by recent attempts to provide explanations from a multitude of perspectives (reviewed in Section 2).

As an alternative approach to deciphering the mysterious neural networks, various network visualization techniques have been actively developed in recent years (e.g. Grün et al. (2016); Seifert et al. (2017) and references therein). Such visualizations not only provide general understanding about the learning process of networks, but also disclose operational instructions on how to adjust network architecture for performance improvements. Majority of visualization approaches probe the relations between input *data* and neuron *activations*, by showing either how neurons *react* to some sample inputs or, reversely, how desired activations are attained or *maximized* with regularized reconstruction of inputs Erhan et al. (2009); Kavukcuoglu et al. (2010); Zeiler et al. (2010); Simonyan et al. (2013); Mahendran & Vedaldi (2015); Yosinski et al. (2015); Alain & Bengio (2016). Input data are invariably used in visualization to probe how the information flow is transformed through the different layers of neural networks. Although insightful, visualization approaches as such have to face a critical open question: to what extend the conclusions drawn from the analysis of sample inputs can be safely applied to new data?

In order to furnish confirmatory answer to the above-mentioned question, ideally, one would have to employ a visualization tool that is *independent* of input data. This ambitious mission appears impossible at a first glance — the final neuron outputs cannot be readily decomposed as the product of *inputs* and neuron *weights* because the thresholding in ReLU activations is input data dependent. By following the principle of fuzzy logic, Fan (2017) recently demonstrated that ReLUs are not essential and can be removed from the so called generalized hamming network (GHN). This simplified network architecture, as reviewed in section 3, facilitates the analysis of neuron interplay based on connection weights only. Consequently, stacked convolution layers can be merged into a single hidden layer without taking into account of inputs from previous layers. Equivalent weights of the merged GHN, which is called *deep epitome*, are computed analytically without resorting to any learning or optimization processes. Moreover, deep epitomes constructed at different layers can be readily applied to new data to extract *hierarchical features* in just *one step* (section 4).

## 2 RELATED WORK

Despite the great success in recent years, neural networks have long been criticized for their black-box natures e.g. in Bentez et al. (1997): "they capture *hidden* relations between inputs and outputs with a highly accurate approximation, but no definitive answer is offered for the question of how they work". The spearhead Mcculloch & Pitts (1943) attempted to interpret neural computing in terms of logic inferencing, followed by more "recent" interpretations e.g. in terms of the *universal approximation* framework Cybenko (1989); Hornik (1991), *restricted Boltzmann machine* Hinton & Salakhutdinov (2006), *information bottleneck* theory Shwartz-Ziv & Tishby (2017), Nevertheless the mission is far from complete and the training of neural networks (especially deep ones) is still a trail-and-error based practice.

The early 1990s witnessed the birth of *fuzzy neural networks* (FNN) Keller et al. (1992); Gupta & Rao (1994) which attempted to furnish neural networks with the interpretability of fuzzy logic Zadeh (1965); Zimmermann (2001); Belohlavek et al. (2017). On the other hand, neural networks have been used as a computational tool to come up with both *membership functions* and fuzzy inference rules Furukawa & Yamakawa (1995); Takagi (2000). This joint force endeavour remains active in the new millennium e.g. Pedrycz & Succi (2002); Liu & Li (2004); Nauck & Nürnberger (2013); Kar et al. (2014); Hu et al. (2016). Nevertheless, FNNs have been largely overlooked nowadays by scholars and engineers in machine learning (ML) community, partially due to the lack of convincing demonstrations on ML problems with large datasets. The exception case is the recent Fan (2017), which re-interpreted celebrated ReLU and batch normalization with a novel Generalized Hamming Network (GHN) and demonstrated the state-of-the-art performances on a variety of machine learning tasks. While GHNs adopted deep networks with multiple convolution layers, in this paper, we will show how to merge multiple stacked convolution layers into a single yet wide convolution layer.

There are abundant empirical evidences backing the belief that deep network structures is preferred to shallow ones Goodfellow et al. (2016), on the other hand, it was theoretically proved by the *universal approximation theorem* that, a single hidden layer network with non-linear activation can well approximate any arbitrary decision functions Cybenko (1989); Hornik (1991). Also, empirically, it was shown that one may reduce depth and increase width of network architecture while still attaining or outperforming the accuracies of deep CNN Ba & Caurana (2013) and residual network Zagoruyko & Komodakis (2016). Nevertheless, it was unclear how to convert a trained deep network into a shallow equivalent network. To this end, the equivalence revealed in Section 3 can be treated as a constructive manifestation of the *universal approximation theorem*.

Various network visualization techniques have been actively developed in recent years, with Erhan et al. (2009) interpreting high level features via maximizing activation and sampling; Kavukcuoglu et al. (2010); Zeiler et al. (2010) learning hierarchical convolutional features via energy or cost minimization; Simonyan et al. (2013) computing class saliency maps for given images; Mahendran & Vedaldi (2015) reconstructing images from CNN features with an natural image prior applied; Yosinski et al. (2015) visualizing live activations as well as deep features via regularized optimization; Alain & Bengio (2016) monitoring prediction errors of individual linear classifiers at multiple iterations. Since all these visualization methods are based on the analysis of examples, the applicability of visualization methods to new data is questionable and no confirmatory answers are provided in a principled manner.

The name "deep epitome" is reminiscent of Jojic et al. (2003); Cheung et al. (2005); Jojic et al. (2010); Chu et al. (2010), in which miniature, condensed "epitomes" consisting of the most essential elements were extracted to *model* and *reconstruct* a set of given images. During the learning process, the self-similarity of image(s), either in terms of pixel-to-pixel comparison or spatial configuration, was exploited and a "smooth" mapping between epitome and input image pixels was estimated.

# 3 DEEP EPITOME

We briefly review generalized hamming networks (GHN) introduced in Fan (2017) and present in great detail a method to derive the deep epitome of a trained GHN. Note that we follow notations in Fan (2017) with minor modifications for the sake of clarity and brevity.

## 3.1 REVIEW OF GHN

According to Fan (2017), the cornerstone notion of generalized hamming distance (GHD) is defined as $g(a,b) := a \oplus b = a + b - 2 \cdot a \cdot b$ for any $a, b \in \mathcal{R}$. Then the negative GHD is used to quantify the similarity between neuron inputs $\mathbf{x}$ and weights $\mathbf{w}$:

$$-g(\mathbf{w}, \mathbf{x}) = \frac{2}{L}\mathbf{w} \cdot \mathbf{x} - \frac{1}{L}\sum_{l=1}^{L} w_l - \frac{1}{L}\sum_{l=1}^{L} x_l, \qquad (1)$$

in which $L$ denotes the length of neuron weights e.g. in convolution kernels, and $g(\mathbf{w}, \mathbf{x})$ is the arithmetic mean of generalized hamming distance between elements of $\mathbf{w}$ and $\mathbf{x}$. By dividing the constant $\frac{2}{L}$, (1) becomes the common representation of neuron computing $(\mathbf{w} \cdot \mathbf{x} + b)$ provided that:

$$b = -\frac{1}{2}\Big(\sum_{l=1}^{L} w_l + \sum_{l=1}^{L} x_l\Big). \qquad (2)$$

It was proposed by Fan (2017) that neuron bias terms should follow the condition (2) analytically without resorting to an optimization approach. Any networks that fulfil this requirement are thus called generalized hamming networks (GHN). In the light of fuzzy logic, the negative of GHD quantifies the *degree of equivalence* between inputs $\mathbf{x}$ and weights $\mathbf{w}$, i.e. the fuzzy truth value of the statement $\mathbf{x} \leftrightarrow \mathbf{w}$ where $\leftrightarrow$ denotes a fuzzy equivalence relation. Moreover, $g(\mathbf{x}, \mathbf{x})$ leads to a measurement of *fuzziness* in $\mathbf{x}$, which reaches the maximal fuzziness when $\mathbf{x} = \mathbf{0.5}$ and monotonically decreases when $\mathbf{x}$ deviates from $\mathbf{0.5}$. Also it can be shown that GHD followed by a non-linear activation induces a fuzzy XOR connective Fan (2017).

When viewed in this GHN framework, the ReLU activation function $max(0, 0.5 - g(\mathbf{x}, \mathbf{w}))$ actually sets a minimal hamming distance threshold of $0.5$ on neuron outputs. Fan (2017) then argued that the use of ReLU activation is not essential because bias terms are analytically set in GHNs. Fan (2017) reported only negligible influences when ReLU was completely skipped for the easy MNIST classification problem. For more challenging CIFAR10/100 classifications, removing ReLUs merely prolonged the learning process but the final classification accuracies remained almost the same. To this end, we restrict our investigation in this paper to those GHNs which have no ReLUs. As illustrated below, this simplification allows for strict derivation of deep epitome from individual convolution layers in GHNs.

## 3.2 GENERALIZED HAMMING DISTANCE AND EPITOME

Fan (2017) postulated that one may analyse the entire GHN in terms of fuzzy logic inference rules, yet no elaboration on the analysis was given. Inspired by the universal approximation framework, we show below how to unravel a deep GHN by merging multiple convolution layers into a single hidden layer.

We first reformulate the convolution operation in terms of *generalized hamming distance* (GHD) for each layer, then illustrate how to combine multiple convolution operations across different layers. As said, this combination is only made possible with GHNs in which bias terms strictly follow condition (2). Without loss of generality, we illustrate derivations and proofs for 1D neuron inputs and weights (with complete proofs elaborated in appendix A). Nevertheless, it is straightforward to extend the derivation to 2D or high dimensions. And appendices B to D illustrate deep epitomes of GHNs trained for 2D MNIST and CIFAR10/100 image classifications.

**Definition 1.** For two given tuples $\mathbf{x}^K = \{x_1, \ldots, x_K\}, \mathbf{y}^L = \{y_1, \ldots, y_L\}$, the *hamming outer product*, denoted $\bigoplus$, is a set of corresponding elements $\mathbf{x}^K \bigoplus \mathbf{y}^L = \{x_k \oplus y_l | k = 1 \ldots K; l = $

Figure 1: **Left panel**: example tuples $X^3, A^2, B^2$; **Middle**: *Hamming outer products* $X^3 \bigoplus A^2$, $X^3 \bigoplus A^2 \bigoplus B^2$; **Right**: *Hamming convolutions* $X^3 \bigoplus^* A^2$, $X^3 \bigoplus^* A^2 \bigoplus^* B^2$ and corresponding *epitomes*. Indices ①,②... ❶,❷... denote subsets $S(1), S(2)\ldots S(n)$ in which element indices satisfying $k + (L-l) = n$ and $k + (L-l) + (M-m) = n$.

$1\ldots L\}$, where $\oplus$ denotes the *generalized hamming distance* operator. Then the product has following properties,

*1. non-commutative*: in general $\mathbf{x}^K \bigoplus \mathbf{y}^L \neq \mathbf{y}^L \bigoplus \mathbf{x}^K$ but they are permutation equivalent, in the sense that there exist permutation matrices $\mathbf{P}$ and $\mathbf{Q}$ such that $\mathbf{x}^K \bigoplus \mathbf{y}^L = \mathbf{P}(\mathbf{y}^L \bigoplus \mathbf{x}^K)\mathbf{Q}$.

*2. non-linear*: in contrast to the standard outer product which is bilinear in each of its entry, the hamming outer product is non-linear since in general $\mathbf{x}^K \bigoplus (\mathbf{y}^L + \mathbf{z}^L) \neq (\mathbf{x}^K \bigoplus \mathbf{y}^L) + (\mathbf{x}^K \bigoplus \mathbf{z}^L)$ and $(\mu\mathbf{x}^K) \bigoplus \mathbf{y}^L \neq \mathbf{x}^K \bigoplus (\mu\mathbf{y}^L) \neq \mu(\mathbf{x}^K \bigoplus \mathbf{y}^L)$ where $\mu \in \mathcal{R}$ is a scalar. Therefore, the hamming outer product defined as such is a *pseudo* outer product.

*3. associative*: $(\mathbf{x}^K \bigoplus \mathbf{y}^L) \bigoplus \mathbf{z}^M = \mathbf{x}^K \bigoplus (\mathbf{y}^L \bigoplus \mathbf{z}^M) = \mathbf{x}^K \bigoplus \mathbf{y}^L \bigoplus \mathbf{z}^M$ because of the associativity of GHD. This property holds for arbitrary number of tuples.

*4. iterated operation*: the definition can be trivially extended to multiple tuples $\mathbf{x}^K \bigoplus \mathbf{y}^L \bigoplus \ldots \mathbf{z}^M = \{x_k \oplus y_l \oplus \ldots, z_m | k = 1\ldots K; l = 1\ldots L; \ldots; m = 1, \ldots M\}$.

**Definition 2.** The *convolution of hamming outer product* or *hamming convolution*, denoted $\bigoplus^*$, of two tuples is a binary operation that sums up corresponding hamming outer product entries:

$$\mathbf{x}^K \overset{*}{\bigoplus} \mathbf{y}^L := \Big\{ \sum_{(k,l) \in S(n)} x_k \oplus y_l \Big| \text{ for } n = 1, \ldots, K+L-1 \Big\} \tag{3}$$

where the subsets $S(n) := \{(k,l) \mid k + (L-l) = n\}$ for $n = 1, \ldots, K+L-1$, and the union of all subsets constitute a partition of all indices $\bigcup_{n=1,\ldots,K+L-1} S(n) = \{(k,l) | k = 1\ldots K; l = 1\ldots L\}$.

The hamming convolution has following properties,

*1. commutative*: $\mathbf{x}^K \bigoplus^* \mathbf{y}^L = \mathbf{y}^L \bigoplus^* \mathbf{x}^K$ since the partition subsets $S(n)$ remains the same.

*2. non-linear*: this property is inherited from the non-linearity of the hamming outer product.

*3. non-associative*: in general $(\mathbf{x}^K \bigoplus^* \mathbf{y}^L) \bigoplus^* \mathbf{z}^M \neq \mathbf{x}^K \bigoplus^* (\mathbf{y}^L \bigoplus^* \mathbf{z}^M)$ since the *summation of GHDs* is non-associative. Note this is in contrast to the associativity of the hamming outer product.

*4. iterated operation*: likewise, the definition can be extended to multiple tuples $\mathbf{x}^K \bigoplus^* \mathbf{y}^L \ldots \mathbf{z}^M = \{ \sum_{(k,l,\ldots,m) \in S(n)} x_k \oplus y_l \ldots \oplus z_m \big| \text{ for } n = 1, \ldots, K+(L-1)+\ldots+(M-1)\}$.

Figure 1 illustrates an example in which GHDs are accumulated through two consecutive convolutions. Note that the conversion from the hamming outer products to its convolution is *non-invertible*, in the sense that, it is impossible to recover individual summands $x_k \oplus y_l$ from the summation $\sum_{(k,l) \in S(n)} x_k \oplus y_l$. As proved in proposition 4, it is possible to compute the convolution of tuples in two (or more) stacked layers without explicitly recovering individual outer product entries of each layer. Due to the non-linearity of the hamming convolutions, computing the composite of two hamming convolutions is non-trivial as elaborated in Section 3.3. In order to illustrate how to carry out this operation, let us first introduce the *epitome* of a hamming convolution as follows.

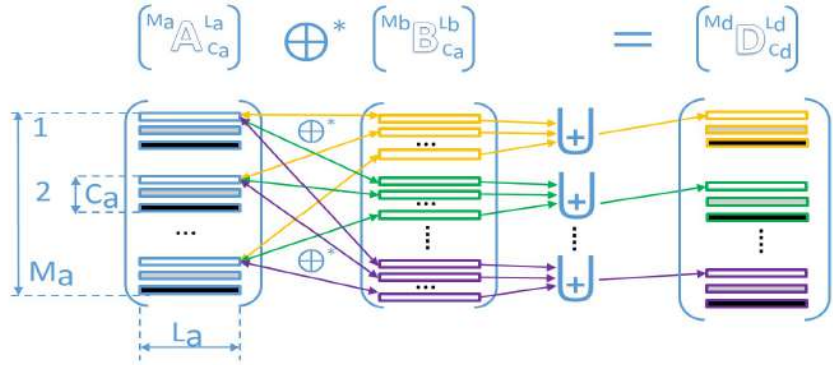

Figure 2: The hamming convolution of two banks of epitomes. Remarks: **a)** for the inputs $\mathbb{A}, \mathbb{B}$ the number of epitomes $M_a$ must be the same as the number of channels $C_b$; and for the output bank $M_d = M_b, C_d = C_a, L_d = (L_a + L_b - 1)$. **b)** the notation $\bigoplus^*$ refers to the hamming convolution between two banks of epitomes (see Definition 5 for details). The convolution of two single-layered epitomes is treated as a special case with all $M_a, C_a, M_b, C_b = 1$. **c)** the notation $\boxplus$ refers to the summation of multiple epitomes of the same length, which is defined in Definition 7. **d)** multiple (coloured) epitomes in $\mathbb{D}$ correspond to different (coloured) epitomes in $\mathbb{B}$; and different (shaded) channels in $\mathbb{D}$ correspond to different (shaded) channels of inputs in $\mathbb{A}$.

**Definition 3.** An *epitome* consists of a set of $N$ pairs $\mathbb{E} = \{(g_n, s_n), |n = 1, \ldots, N\}$ where $g_n$ denotes the summation of GHD entries from some hamming convolutions, $s_n$ the number of summands or the cardinality of the subset $S(n)$ defined above, and $N$ is called the *length* of the epitome.

A *normalized epitome* is an epitome with $s_n = 1$ for all $n = 1, \ldots N$. Any epitome can then be *normalized* by setting $(g_n/s_n, 1)$ for all elements. A normalized epitome may also refer to input data $\mathbf{x}$ or neuron weights $\mathbf{w}$ that are not yet involved in any convolution operations. In the latter case, $g_n$ is simply the input data $\mathbf{x}$ or neuron weights $\mathbf{w}$.

*Remark*: the summation of GHD entries $g_n$ is defined abstractly, and depending on different scenarios, the underlying outer product may operate on arbitrary number of tuples $g_n = \left(\mathbf{x}^K \bigoplus^* \mathbf{y}^L \ldots \mathbf{z}^M\right)(n) = \sum_{(k,l,\ldots,m) \in S(n)} x_k \oplus y_l \ldots \oplus z_m$.

**Fuzzy logic interpretation**: in contrast to the traditional signal processing point of view, in which neuron weights $\mathbf{w}$ are treated as parameters of linear transformation and bias terms $b$ are appropriate thresholds for non-linear activations, the generalized hamming distance approach *treats $\mathbf{w}$ as fuzzy templates and sets bias terms analytically* according to (2). In this view, the normalization $g_n/s_n$ is nothing but the *mean GHD* of entries in the subset $S(n)$, which indicates a *grade of fitness* (or a fuzzy set) between templates $\mathbf{w}$ and inputs $\mathbf{x}$ at location $n$. This kind of arithmetic mean operator has been used for aggregating evidences in fuzzy sets and empirically performed quite well in decision making environments (e.g. see Zimmermann (2001)).

Still in the light of signal processing, the generalized hamming distance naturally induces an *information enhancement and suppression* mechanism. Since the gradient of $g(\mathbf{x}, \mathbf{w})$ with respect to $\mathbf{x}$ is $1 - 2\mathbf{w}$, the information in $\mathbf{x}$ is then either enhanced or suppressed according to $\mathbf{w}$ : a) the output $g(\mathbf{x}, \mathbf{w})$ is always $\mathbf{x}$ for $\mathbf{w} = 0$ (conversely $1 - \mathbf{x}$ for $\mathbf{w} = 1$) with no information loss in $\mathbf{x}$; b) for $\mathbf{w} = 0.5$, the output $g(\mathbf{x}, \mathbf{w})$ is always 0.5 regardless of $\mathbf{x}$, thus input information in $\mathbf{x}$ is completely suppressed; c) for $\mathbf{w} < 0.0$ or $\mathbf{w} > 1.0$ information in $\mathbf{x}$ is proportionally enhanced. It was indeed observed, during the learning process in our experiments, a small faction of prominent feature pixels in weights **w** gradually attain large positive or negative values, so that corresponding input pixels play decisive roles in classification. On the other hand, large majority of obscure pixels remain in the fuzzy regime near 0.5, and correspondingly, input pixels have virtually no influence on the final decision (see experimental results in Section 4). This observation is also in accordance with the *information compression* interpretation advocated by Shwartz-Ziv & Tishby (2017), and the connection indicates an interesting research direction for future work.

### 3.3 DEEP EPITOME

This subsection only illustrates main results concerning how to merge multiple hamming convolution operations in stacked layers into a single-layer of epitomes i.e. *deep epitome*. Detailed proofs are given in appendix A.

**Notation**: for the sake of brevity, let $[^{M_a}\mathbb{A}_{C_a}^{L_a}]$ denote a *bank of epitomes*: $\{^m\mathbb{A}_{C_a}^{L_a} \mid m = 1, \ldots M_a\}$, where $\mathbb{A}_{C_a}^{L_a} = \{\mathbb{A}_c^{L_a} \mid c = 1, \ldots C_a\}$ are $C_a$-channels of length-$L_a$ epitomes, and $M_a$ is the number of epitomes as such in the bank or set $[\mathbb{A}]$. Figure 2 illustrates example banks of epitomes and two operations defined on them (also see Appendix A for detailed definition of $\uplus$).

**Theorem 10.** *A generalized hamming network consisting of multiple convolution layers, is equivalent to a bank of epitome, called deep epitome $[^\star\mathbb{D}_\nabla^\diamond]$, which can be computed by iteratively applying the composite hamming convolution in equation (8) to individual layer of epitomes:*

$$[^\star\mathbb{D}_\nabla^\diamond] := [^{M_a}\mathbb{A}_{C_a}^{L_a}] \overset{*}{\bigoplus} [^{M_b}\mathbb{B}_{C_b}^{L_b}] \overset{*}{\bigoplus} \ldots \overset{*}{\bigoplus} [^{M_z}\mathbb{Z}_{C_z}^{L_z}], \quad (10)$$

*in which $\nabla = C_a$ is the number of channels in the first bank $\mathbb{A}$, $\star = M_z$ is the number of epitomes in the last bank $\mathbb{Z}$, and $\diamond = L_a + (L_b - 1) + \ldots + (L_z - 1)$ is the length of composite deep epitome. Note that for the hamming convolution to be a valid operation, the number of epitomes in the previous layer and the number channels in the current layer must be the same e.g. $C_b = M_a$.*

*Proof.* For given inputs represented as a bank of normalized epitomes $[^{M_x}\mathbb{X}_{C_x}^{L_x}]$ the final network output $[^{M_z}Y_{C_x}^{L_y}]$ is obtained by recursively applying equation (8) to outputs from the previous layers, and factoring out the input due to the associativity proved in proposition 9:

$$[^{M_z}Y_{C_x}^{L_y}] = \left(\left(\left([^{M_x}\mathbb{X}_{C_x}^{L_x}] \overset{*}{\bigoplus} [^{M_a}\mathbb{A}_{C_a}^{L_a}]\right) \overset{*}{\bigoplus} [^{M_b}\mathbb{B}_{C_b}^{L_b}]\right) \overset{*}{\bigoplus} \ldots \overset{*}{\bigoplus} [^{M_z}\mathbb{Z}_{C_z}^{L_c}]\right)$$
$$= [^{M_x}\mathbb{X}_{C_x}^{L_x}] \overset{*}{\bigoplus} \underbrace{\left([^{M_a}\mathbb{A}_{C_a}^{L_a}] \overset{*}{\bigoplus} [^{M_b}\mathbb{B}_{C_b}^{L_b}] \overset{*}{\bigoplus} \ldots \overset{*}{\bigoplus} [^{M_z}\mathbb{Z}_{C_z}^{L_c}]\right)}_{[^\star\mathbb{D}_\nabla^\diamond]}. \quad (11)$$

□

Remark: due to the non-linearity of underlying hamming outer products, to prove the associativity of the *convolution of epitomes* is by no means trivial (see proposition 9). In essence, we have to use proposition 4 to compute the convolution of two epitomes even though individual entries of the underlying hamming outer product are not directly accessible. Consequently, the updating rule outlined in equations (4) and (5) play the crucial role in setting due bias terms analytically for generalized hamming networks (GHN), as opposed to the optimization approach often adopted by many non-GHN deep convolution networks.

**Fuzzy logic inferencing with deep epitomes**: Eq. (11) can be treated as a fuzzy logic inferencing rule, with which elements of input **x** are compared with respect to corresponding elements of deep epitomes **d**. More specifically, the negative of GHD quantifies the *degree of equivalence* between inputs **x** and epitome weights **d**, i.e. the fuzzy truth value of the assertion $\mathbf{x} \leftrightarrow \mathbf{d}$ where $\leftrightarrow$ denotes a fuzzy *logical biconditional*. Therefore, output scores in **y** indicate the grade of fuzzy equivalences truth values between **x** and the shifted **d** at different spatial locations. This inferencing rule, in the same vein of Fan (2017), is applicable to either a single layer neuron weights or the composite deep epitomes as proved by (11).

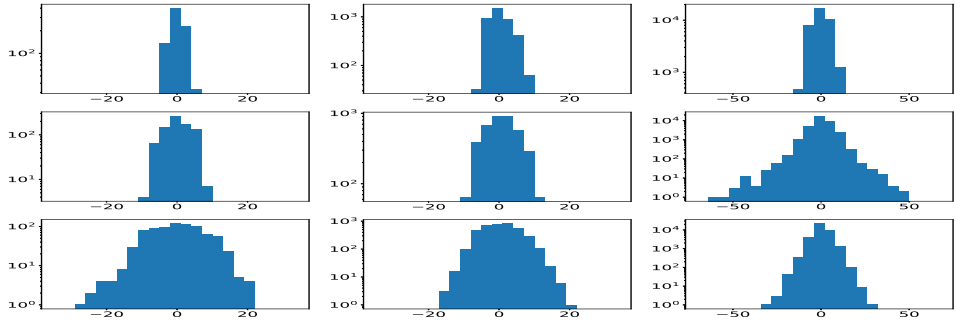

Figure 3: Histograms of *normalized* deep epitomes at different layers/iterations for GHN trained with MNIST classification. Left to right: layers 1,2,3. Top, middle and bottom rows: iteration 200, 1000 and 10000 respectively.

**Constructive manifestation of the universal approximation theorem**: it was proved that a single hidden layer network with non-linear activation can well approximate any arbitrary decision functions Cybenko (1989); Hornik (1991), yet it was also argued by Goodfellow et al. (2016) that such a single layer may be infeasibly large and may fail to learn and generalize correctly. Theorem 10 proves that such a simplified single hidden layer network can actually be constructed from a trained GNH. In this sense Theorem 10 illustrates a concrete solution which materializes the universal approximation theorem.

## 4 DEEP EPITOME FOR NETWORK VISUALIZATION

We illustrate below deep epitomes extracted from three generalized hamming networks trained with MNIST, CIFAR10/100 classification respectively. Detailed descriptions about the network architectures (number of layers, channels etc.) are included in the appendix.

### 4.1 DATA INDEPENDENT VISUALIZATION OF DEEP EPITOMES

Deep epitomes derived in the previous section allows one to build up and visualize hierarchical features in an on-line manner during the learning process. This approach is in contrast to many existing approaches, which often apply additional optimization or learning processes with various type of regularizations e.g. in Erhan et al. (2009); Zeiler et al. (2010); Simonyan et al. (2013); Mahendran & Vedaldi (2015); Yosinski et al. (2015). Figures 5, 8 and 11, 12 in appendices illustrate deep epitomes learnt by three generalized hamming networks for the MNIST and CIFAR10/100 image classification tasks. It was observed that geometrical structures of hierarchical features were formed at different layers, rather early during the learning process (e.g. 1000 out of 10000 iterations). Substantial follow up efforts were invested on refining features for improved details. The scrutinization of normalized epitome histograms in Figure 3 showed that a majority of pixel values remain relatively small during the learning process, while a small fraction of epitome weights gradually accumulate large values over thousands of iterations to form prominent features.

The observation of sparse features has been reported and interpreted in terms of sparse coding e.g. Papyan et al. (2016) or the information compression mechanism as advocated by Shwartz-Ziv & Tishby (2017). Following Fan (2017) we adopt the notion of *fuzziness* (also reviewed in Section 3.1) to provide a fuzzy logic interpretation: *prominent features correspond to neuron weights with low fuzziness*. It was indeed observed in Figure 4 that *fuzziness* of deep epitomes in general decrease during the learning process despite of fluctuations at some layers. The inclination towards *reduced fuzziness* seems in accord with the minimization of classification errors, although the fuzziness is not explicitly minimized.

Finally we re-iterate that the internal representation of deep epitomes is *input data independent*. For instance in MNIST handwritten images, it is certain constellations of strokes instead of digits that are learnt at layer 3 (see Figure 5). The matching of arbitrary input data with such "fuzzy templates" is then quantified by the generalized hamming distance, and can be treated as generic fuzzy logic inferencing rules learnt by GHNs. The matching score measured by GHDs can also be treated as salient features that are subsequently fed to the next layer (see Section 4.2 with Figures 6 and more results in appendices B and C)[1].

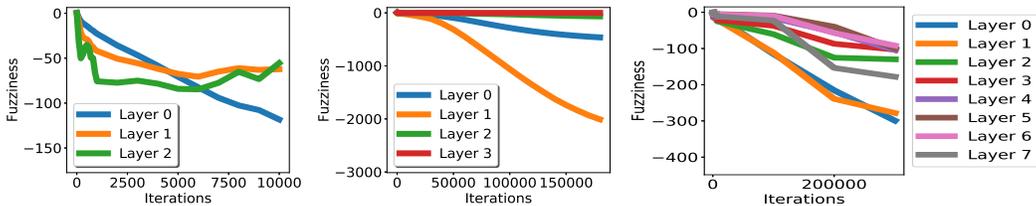

Figure 4: Fuzziness in *normalized* deep epitomes at different layers and learning iterations. **Left**: a GHN trained with MNIST classification. **Middle**: CIFAR10. **Right**: CIFAR100. See section 3.1 for definition of *fuzziness*.

### 4.2 DATA DEPENDENT FEATURE EXTRACTION

Feature extraction of given inputs is straightforward with deep epitomes applied according to eq. (11). Figures 6 (and more results in appendices B and C) show example features extracted at different layers of GHN trained on MNIST, CIFAR10/100 image datasets. Clearly extracted features represent different types of salient features e.g. oriented strokes in hand written images, oriented edgelets, textons with associated colours or even rough segmentations in CIFAR images. These features all become gradually more discriminative during the learning process.

It must be noted that the extraction of these hierarchical salient features is not entirely new and has been reported e.g. in Erhan et al. (2009); Kavukcuoglu et al. (2010). Nevertheless, the equivalence of deep epitomes disclosed in Theorem 10 leads to an unique characteristic of GHNs — deep layer features do not necessarily rely on features extracted from previous layers, instead, they can be extracted in *one step* using deep epitomes at desired layers. For extremely deep convolution networks e.g. those with over 100 layers, this simplification may bring about substantial reduction of computational and algorithmic complexities. This potential advantage is worth follow up exploration in future research.

## 5 DISCUSSION AND CONCLUSION

We have proposed in this paper a novel network representation, called *deep epitome*, which is proved to be equivalent to stacked convolution layers in generalized hamming networks (GHN). Theoretically this representation provides a constructive manifestation for the *universal approximation theorem* Cybenko (1989); Hornik (1991), which states that a single layered network, in principle, is able to approximate any arbitrary decision functions up to any desired accuracy. On the other hand, it is a dominant belief Goodfellow et al. (2016), which is supported by abundant empirical evidences, that deep structures play an indispensable role in decomposing the combinatorial optimization problem into layer-wise manageable sub-problems. We concur with the view and supplement with our demonstration that, a trained deep GHN can be converted into a simplified networks for the sake of high interpretability, reduced algorithmic and computational complexities.

The success of our endeavours lies in the rigorous derivation of convolving epitomes across different layers in eq. (4) and (5), which set due bias terms analytically without resorting to optimization-based approaches. Consequently, deep epitomes at all convolution layers can be computed without using any input data. Moreover, deep epitomes can be used to extract hierarchical features in just *one step* at any desired layers. In the light of fuzzy logic, the normalized epitome (definition 3) encodes a *grade of fitness* between the learnt templates and given inputs at certain spatial locations. This fuzzy logic interpretation furnishes a refreshing perspective that, in our view, will open the black box of deep learning eventually.

---

[1]This learnt "fuzzy template" is reminiscent of epitomes in Jojic et al. (2003) and gives rise to the name *deep epitome*.

## APPENDIX A

**Definition 1.** For two given tuples $\mathbf{x}^K = \{x_1, \ldots, x_K\}, \mathbf{y}^L = \{y_1, \ldots, y_L\}$, the *hamming outer product*, denoted $\bigoplus$, is a set of corresponding elements $\mathbf{x}^K \bigoplus \mathbf{y}^L = \{x_k \oplus y_l | k = 1 \ldots K; l = 1 \ldots L\}$, where $\oplus$ denotes the *generalized hamming distance* operator. Then the product has following properties,

*1. non-commutative*: in general $\mathbf{x}^K \bigoplus \mathbf{y}^L \neq \mathbf{y}^L \bigoplus \mathbf{x}^K$ but they are permutation equivalent, in the sense that there exist permutation matrices $\mathbf{P}$ and $\mathbf{Q}$ such that $\mathbf{x}^K \bigoplus \mathbf{y}^L = \mathbf{P}(\mathbf{y}^L \bigoplus \mathbf{x}^K)\mathbf{Q}$.

*2. non-linear*: in contrast to the standard outer product which is bilinear in each of its entry, the hamming outer product is non-linear since in general $\mathbf{x}^K \bigoplus (\mathbf{y}^L + \mathbf{z}^L) \neq (\mathbf{x}^K \bigoplus \mathbf{y}^L) + (\mathbf{x}^K \bigoplus \mathbf{z}^L)$ and $(\mu \mathbf{x}^K) \bigoplus \mathbf{y}^L \neq \mathbf{x}^K \bigoplus (\mu \mathbf{y}^L) \neq \mu(\mathbf{x}^K \bigoplus \mathbf{y}^L)$ where $\mu \in \mathcal{R}$ is a scalar. Therefore, the hamming outer product defined as such is a *pseudo* outer product.

*3. associative*: $(\mathbf{x}^K \bigoplus \mathbf{y}^L) \bigoplus \mathbf{z}^M = \mathbf{x}^K \bigoplus (\mathbf{y}^L \bigoplus \mathbf{z}^M) = \mathbf{x}^K \bigoplus \mathbf{y}^L \bigoplus \mathbf{z}^M$ because of the associativity of GHD. This property holds for arbitrary number of tuples.

*4. iterated operation*: the definition can be trivially extended to multiple tuples $\mathbf{x}^K \bigoplus \mathbf{y}^L \bigoplus \ldots \mathbf{z}^M = \{x_k \oplus y_l \oplus \ldots, z_m | k = 1 \ldots K; l = 1 \ldots L; \ldots; m = 1, \ldots M\}$.

*Proof. associativity*: by definition it suffices to prove element-wise $(x_k \oplus y_l) \oplus z_m = x_k \oplus (y_l \oplus z_m)$ because of the associativity of the generalized hamming distance.

*non-linearity*: by definition $\mathbf{x}^K \bigoplus (\mathbf{y}^L + \mathbf{z}^L)$ has elements $x_k \oplus (y_l + z_l)$, then it suffices to prove non-linearity for each element i.e. $x_k \oplus (y_l + z_l) = x_k + (y_l + z_l) - 2x_k(y_l + z_l) \neq (x_k + y_l - 2x_k y_l) + (x_k + z_l - 2x_k z_l) = (x_k \oplus y_l) + (x_k \oplus z_l)$. Similarly, $(\mu x_k \oplus y_l) = \mu x_k + y_l - 2\mu x_k y_l \neq x_k + \mu y_l - 2\mu x_k y_l \neq \mu(x_k + y_l - 2x_k y_l)$ in general. □

**Definition 2.** The *convolution of hamming outer product* or *hamming convolution*, denoted $\bigoplus^*$, of two tuples is a binary operation that sums up corresponding hamming outer product entries:

$$\mathbf{x}^K \overset{*}{\bigoplus} \mathbf{y}^L := \Big\{ \sum_{(k,l) \in S(n)} x_k \oplus y_l \,\Big|\, \text{for } n = 1, \ldots, K + L - 1 \Big\} \tag{3}$$

where the subsets $S(n) := \{(k, l) \mid k + (L - l) = n\}$ for $n = 1, \ldots, K + L - 1$, and the union of all subsets constitute a partition of all indices $\bigcup_{n=1,\ldots,K+L-1} S(n) = \{(k,l) | k = 1 \ldots K; l = 1 \ldots L\}$.

The hamming convolution has following properties,

*1. commutative*: $\mathbf{x}^K \bigoplus^* \mathbf{y}^L = \mathbf{y}^L \bigoplus^* \mathbf{x}^K$ since the partition subsets $S(n)$ remains the same.

*2. non-linear*: this property is inherited from the non-linearity of the hamming outer product.

*3. non-associative*: in general $(\mathbf{x}^K \bigoplus^* \mathbf{y}^L) \bigoplus^* \mathbf{z}^M \neq \mathbf{x}^K \bigoplus^* (\mathbf{y}^L \bigoplus^* \mathbf{z}^M)$ since the *summation of GHDs* is non-associative. Note this is in contrast to the associativity of the hamming outer product.

*4. iterated operation*: likewise, the definition can be extended to multiple tuples $\mathbf{x}^K \bigoplus^* \mathbf{y}^L \ldots \mathbf{z}^M = \{\sum_{(k,l,\ldots,m) \in S(n)} x_k \oplus y_l \ldots \oplus z_m \,|\, \text{for } n = 1, \ldots, K + (L-1) + \ldots + (M-1)\}$.

*Proof. non-associativity*: by definition it suffices to prove element-wise in general
$\sum_{(n,m) \in S'(n')} \big( \sum_{(k,l) \in S(n)} x_k \oplus y_l \big) \oplus z_m \neq \sum_{(k,n) \in S'(n')} x_k \oplus \big( \sum_{(l,m) \in S(n)} y_l \oplus z_m \big)$. □

**Definition 3.** An *epitome* consists of a set of $N$ pairs $\mathbb{E} = \{(g_n, s_n), | n = 1, \ldots, N\}$ where $g_n$ denotes the summation of GHD entries from some hamming convolutions, $s_n$ the number of summands or the cardinality of the subset $S(n)$ defined above, and $N$ is called the *length* of the epitome.

A *normalized epitome* is an epitome with $s_n = 1$ for all $n = 1, \ldots N$. Any epitome can then be *normalized* by setting $(g_n/s_n, 1)$ for all elements. A normalized epitome may also refer to input data $\mathbf{x}$ or neuron weights $\mathbf{w}$ that are not yet involved in any convolution operations. In the latter case, $g_n$ is simply the input data $\mathbf{x}$ or neuron weights $\mathbf{w}$.

**Proposition 4.** *Given two tuples* $\mathbf{x} = \{x_k | k = 1 \ldots K\}$ *and* $\mathbf{y} = \{y_l | l = 1 \ldots L\}$, *then*

$$\sum_k^K \sum_l^L (x_k \oplus y_l) = \Big(\sum_k^K x_k\Big) \oplus \Big(\sum_l^L y_l\Big) + (L-1)\sum_k^K x_k + (K-1)\sum_l^L y_l. \tag{4}$$

*Proof.*

$$LHS = \sum_k^K \sum_l^L (x_k + y_l - 2x_k y_l) = L\sum_k^K x_k + K\sum_l^L y_l - 2\sum_k^K x_k \sum_l^L y_l$$

$$= \Big(\sum_k^K x_k + \sum_l^L y_l - 2\sum_k^K x_k \sum_l^L y_l\Big) + (L-1)\sum_k^K x_k + (K-1)\sum_l^L y_l$$

$$= \Big(\sum_k^K x_k\Big) \oplus \Big(\sum_l^L y_l\Big) + (L-1)\sum_k^K x_k + (K-1)\sum_l^L y_l = RHS$$

□

Remark: eq. (4) allows one to compute summation of all hamming outer product elements on the right hand side, even though individual elements $x_k$ and $y_l$ are unable to recover from the given summands $\sum_k x_k$ and $\sum_l y_l$. The definition below immediately follows and illustrates how to merge elements of two epitomes.

**Definition 5.** Given two epitomes $\mathbb{E}_a = \{(g_n, s_n) | n = 1, \ldots N\}, \mathbb{E}_b = \{(g'_m, s'_m) | m = 1, \ldots M\}$, the *convolution of two epitomes* $\mathbb{E}_c = \mathbb{E}_a \bigoplus^* \mathbb{E}_b$ is given by:

$$\mathbb{E}_c = \{(g''_c, s''_c) | c = 1, \ldots, N + M - 1\}; \tag{5a}$$

$$\text{where } g''_c = \sum_{(n,m) \in S(c)} \Big( g_n \oplus g'_m + (s'_m - 1)g_n + (s_n - 1)g'_m \Big), \tag{5b}$$

$$s''_c = \sum_{(n,m) \in S(c)} s_n s'_m, \tag{5c}$$

$$S(c) := \{(n,m) | n + (M - m) = c\}. \tag{5d}$$

*Proof.* For each pair of epitome elements $(g_n, s_n)$ and $(g'_m, s'_m)$ since by definition 3 $g_n = \sum_{k=1}^{s_n} x_k$ is a summation of elements and $g'_m$ in the same vein, then the summation of hamming outer product elements $\sum_{k=1}^{s_n} \sum_{l=1}^{s'_m} (x_k \oplus y_l)$ follows eq. (4). The number of elements $s''_c$ is simply the convolution of $s_n$ and $s'_m$ of two given epitomes. □

Remark: this operation is applicable to the case when two epitomes are merged via *spatial convolution* (see Figure 2 for an example). Note that this merging operation is *associative* due to the following theorem.

**Theorem 6.** *The convolution of multiple epitomes, as defined in 5, is associative:*

$$\mathbb{E}_a \bigoplus^* \mathbb{E}_b \bigoplus^* \mathbb{E}_c = (\mathbb{E}_a \bigoplus^* \mathbb{E}_b) \bigoplus^* \mathbb{E}_c = \mathbb{E}_a \bigoplus^* (\mathbb{E}_b \bigoplus^* \mathbb{E}_c). \tag{6}$$

*Proof.* By definition 5, elements of $\mathbb{E}_a \bigoplus^* \mathbb{E}_b$ are the summations of hamming outer product elements denoted by $\sum_{k=1}^{s_n} \sum_{l=1}^{s'_m} (x_k \oplus y_l)$. Then elements of $(\mathbb{E}_a \bigoplus^* \mathbb{E}_b) \bigoplus^* \mathbb{E}_c$ are the summation of hamming outer product elements $\sum_{k=1}^{s_n} \sum_{l=1}^{s'_m} \sum_{i=1}^{s''_q} (x_k \oplus y_l \oplus z_i)$, which are equal to elements of $\mathbb{E}_a \bigoplus^* (\mathbb{E}_b \bigoplus^* \mathbb{E}_c)$, due to the associativity of hamming outer products by definition 1.

□

**Remark**: this associative property is of paramount importance for the derivation of deep epitomes, which factor out the inputs **x** from subsequent convolutions with neuron weights **w**.

**Definition 7.** Given two epitomes of the same size $\mathbb{E}_a = \{(g_n, s_n) | n = 1, \ldots N\}, \mathbb{E}_b = \{(g'_n, s'_n) | n = 1, \ldots N\}$, the *summation of two epitomes* $\mathbb{E}_c = \mathbb{E}_a \uplus \mathbb{E}_b$ is trivially defined by element-wise summation:

$$\mathbb{E}_c = \{(g''_n, s''_n) | n = 1, \ldots, N\};$$
$$\text{where } g''_n = g_n + g'_n, \quad (7)$$
$$s''_n = s_n + s'_n.$$

Remark: the summation operation is applicable to the case when epitomes are (iteratively) merged cross different channels (see Figure 2 for an example). Note that the size of two input epitomes must be the same, and the size of output epitome remain unchanged. Moreover, the operation is trivially extended to multiple epitomes $\underset{\{1,2,\ldots,M\}}{\uplus} := \mathbb{E}_1 \uplus \mathbb{E}_2 \uplus \ldots \uplus \mathbb{E}_M$.

**Notation**: for the sake of brevity, let $[^{M_a}\mathbb{A}_{C_a}^{L_a}]$ denote a *bank of epitomes*: $\{^m\mathbb{A}_{C_a}^{L_a} \mid m = 1, \ldots M_a\}$, where $\mathbb{A}_{C_a}^{L_a} = \{\mathbb{A}_c^{L_a} \mid c = 1, \ldots C_a\}$ are $C_a$-channels of length-$L_a$ epitomes, and $M_a$ is the number of epitomes as such in the bank or set $[\mathbb{A}]$. Figure 2 illustrates example banks of epitomes and two operations defined on them.

**Definition 8.** The *composite convolution* of two banks of epitomes $[^{M_a}\mathbb{A}_{C_a}^{L_a}]$ and $[^{M_b}\mathbb{B}_{C_b}^{L_b}]$ with $M_a = C_b$, is defined as

$$[^{M_a}\mathbb{A}_{C_a}^{L_a}]\overset{*}{\bigoplus}[^{M_b}\mathbb{B}_{C_b}^{L_b}] := \left\{ \underset{m_a=1, c_b=1}{\overset{M_a, C_b}{\uplus}} \left( ^{m_a}\mathbb{A}_{c_a}^{L_a} \overset{*}{\bigoplus} {}^{m_b}\mathbb{B}_{c_b}^{L_b} \right) \middle| c_a = 1, \ldots C_a; m_b = 1, \ldots M_b \right\}. \quad (8)$$

The output of this operation, in turn, is a bank with $M_b$ of $C_a$-channel length-$(L_a + L_b - 1)$ epitomes denoted as $[^{M_d}\mathbb{D}_{C_d}^{L_d}]$ with $M_d = M_b, C_d = C_a, L_d = L_a + L_b - 1$. See Figure 2 for an example.

**Proposition 9.** *The composite convolutions of multiple epitome banks, as given in definition 8, is associative:*

$$\left([^{M_a}\mathbb{A}_{C_a}^{L_a}]\overset{*}{\bigoplus}[^{M_b}\mathbb{B}_{C_b}^{L_b}]\right)\overset{*}{\bigoplus}[^{M_c}\mathbb{C}_{C_c}^{L_c}] = [^{M_a}\mathbb{A}_{C_a}^{L_a}]\overset{*}{\bigoplus}\left([^{M_b}\mathbb{B}_{C_b}^{L_b}]\overset{*}{\bigoplus}[^{M_c}\mathbb{C}_{C_c}^{L_c}]\right) \quad (9)$$

*Proof.* The associativity immediately follows the associativity of Theorem 6 and definition 7. □

Remark: this associative property, which is inherited from theorem 6, can be trivially extended to multiple banks and lead to the main theorem of the paper as follows.

**Theorem 10.** *A generalized hamming network consisting of multiple convolution layers, is equivalent to a bank of epitome, called deep epitome $[^\star \mathbb{D}_\triangledown^\diamond]$, which can be computed by iteratively applying the composite hamming convolution in equation (8) to individual layer of epitomes:*

$$[^\star \mathbb{D}_\triangledown^\diamond] := [^{M_a}\mathbb{A}_{C_a}^{L_a}]\overset{*}{\bigoplus}[^{M_b}\mathbb{B}_{C_b}^{L_b}]\overset{*}{\bigoplus}\ldots\overset{*}{\bigoplus}[^{M_z}\mathbb{Z}_{C_z}^{L_z}], \quad (10)$$

*in which $\triangledown = C_a$ is the number of channels in the first bank $\mathbb{A}$, $\star = M_z$ is the number of epitomes in the last bank $\mathbb{Z}$, and $\diamond = L_a + (L_b - 1) + \ldots + (L_z - 1)$ is the length of composite deep epitome. Note that for the hamming convolution to be a valid operation, the number of epitomes in the previous layer and the number channels in the current layer must be the same e.g. $C_b = M_a$.*

*Proof.* For given inputs represented as a bank of normalized epitomes $[^{M_x}\mathbb{X}_{C_x}^{L_x}]$ the final network output $[^{M_z}Y_{C_x}^{L_y}]$ is obtained by recursively applying equation (8) to outputs from the previous layers, and factoring out the input due to the associativity proved in proposition 9:

$$\begin{aligned}[^{M_z}Y^{L_y}_{C_x}] &= \left(\left(\left([^{M_x}\mathbb{X}^{L_x}_{C_x}]\overset{*}{\bigoplus}[^{M_a}\mathbb{A}^{L_a}_{C_a}]\right)\overset{*}{\bigoplus}[^{M_b}\mathbb{B}^{L_b}_{C_b}]\right)\overset{*}{\bigoplus}\cdots\overset{*}{\bigoplus}[^{M_z}\mathbb{Z}^{L_c}_{C_z}]\right)\\ &= [^{M_x}\mathbb{X}^{L_x}_{C_x}]\overset{*}{\bigoplus}\underbrace{\left([^{M_a}\mathbb{A}^{L_a}_{C_a}]\overset{*}{\bigoplus}[^{M_b}\mathbb{B}^{L_b}_{C_b}]\overset{*}{\bigoplus}\cdots\overset{*}{\bigoplus}[^{M_z}\mathbb{Z}^{L_c}_{C_z}]\right)}_{[{}^\star\mathbb{D}^\diamond_\triangledown]}.\end{aligned} \quad (11)$$

□

**Network architectures used in Appendices B,C,D**:

We summarize in Tables below architectures of three generalized hamming networks trained with MNIST, CIFAR10/100 classification respectively. Note that for kernels with stride 2, we resize original kernels to their effective size (×2) when computing deep epitomes. Also we use average-pooling, instead of max-pooling, in all three networks. Subsequent fully connected layers are also reported although they are not involved in the computation of deep epitomes.

| GHN for MNIST classification | | | | | |
|---|---|---|---|---|---|
| Layers | Kernel | Str. | resized | Ch I/O | **Epitome size** |
| conv1 | 5x5 | 1 | 5x5 | 3 / 32 | 5x5 |
| conv2 | 5x5 | 2 | 10x10 | 32 / 32 | 14x14 |
| conv3 | 5x5 | 2 | 10x10 | 32/ 128 | 23x23 |
| fc1 | - | - | - | * / 1024 | - |
| fc2 | - | - | - | 1024 / 10 | - |

| GHN for CIFAR10 classification | | | | | |
|---|---|---|---|---|---|
| Layers | Kernel | Str. | resized | Ch I/O | **Epitome size** |
| conv1 | 3x3 | 1 | 3x3 | 3 / 64 | 3x3 |
| conv2 | 3x3 | 1 | 3x3 | 64 / 64 | 5x5 |
| conv3 | 5x5 | 2 | 10x10 | 64 / 256 | 14x14 |
| conv4 | 5x5 | 2 | 10x10 | 256 / 256 | 23x23 |
| fc1 | - | - | - | */1024 | - |
| fc2 | - | - | - | 1024/512 | - |
| fc3 | - | - | - | 512/10 | - |

| GHN for CIFAR100 classification | | | | | |
|---|---|---|---|---|---|
| Layers | Kernel | Str. | resized | Ch I/O | **Epitome size** |
| conv1 | 3x3 | 1 | 3x3 | 3 / 64 | 3x3 |
| conv2 | 5x5 | 2 | 10x10 | 64 / 64 | 12x12 |
| conv3 | 5x5 | 1 | 5x5 | 64 / 64 | 16x16 |
| conv4 | 5x5 | 1 | 5x5 | 64 / 64 | 20x20 |
| conv5 | 5x5 | 1 | 5x5 | 64 / 64 | 24x24 |
| conv6 | 5x5 | 1 | 5x5 | 64 / 64 | 28x28 |
| conv7 | 5x5 | 2 | 10x10 | 64 / 128 | 37x37 |
| fc1 | - | - | - | */1024 | - |
| fc2 | - | - | - | 1024/512 | - |
| fc3 | - | - | - | 512/10 | - |

APPENDIX B: DEEP EPITOMES WITH MNIST HANDWRITTEN RECOGNITION

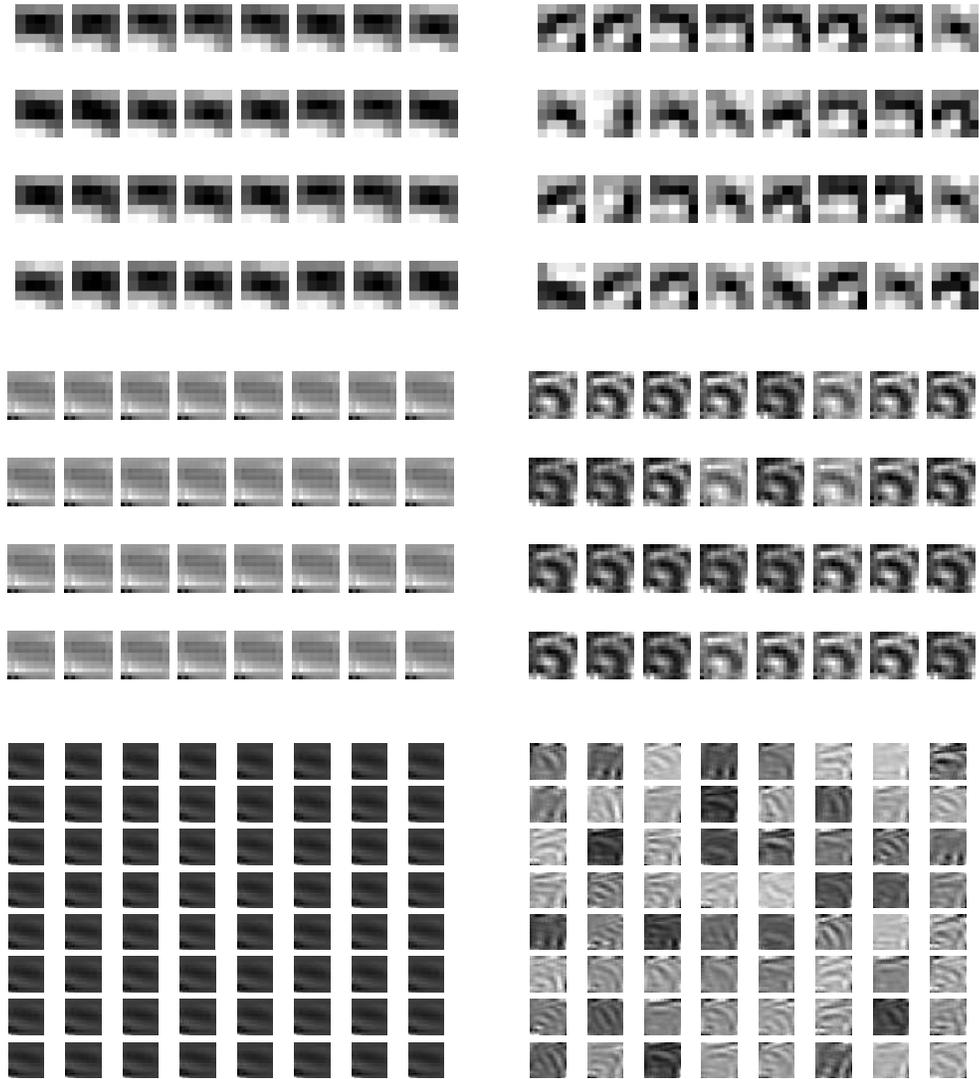

Figure 5: Deep epitomes at layers 1,2 and 3 for a GHN trained with MNIST classification at iterations 100 and 10000 respectively.

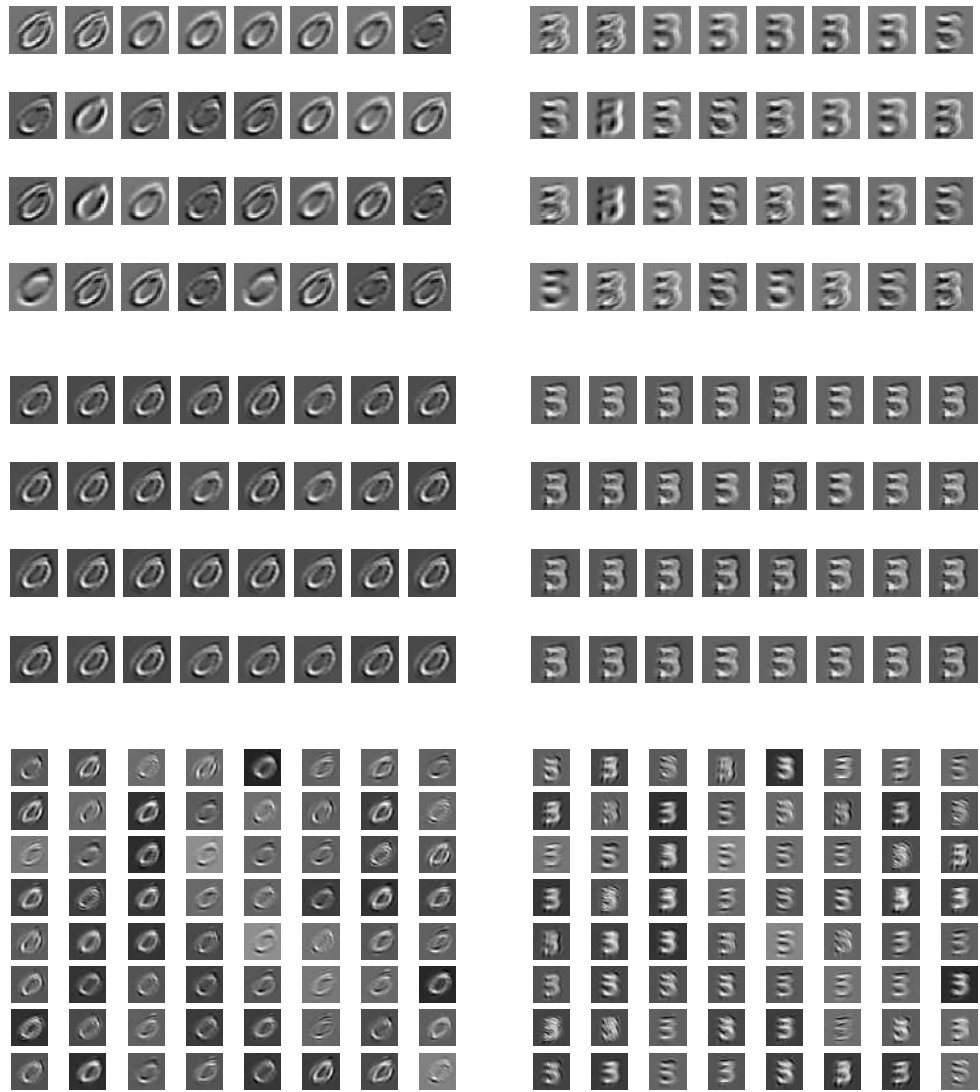

Figure 6: Example hierarchical features extracted at layers 1,2, and 3 for a GHN trained with MNIST classification.

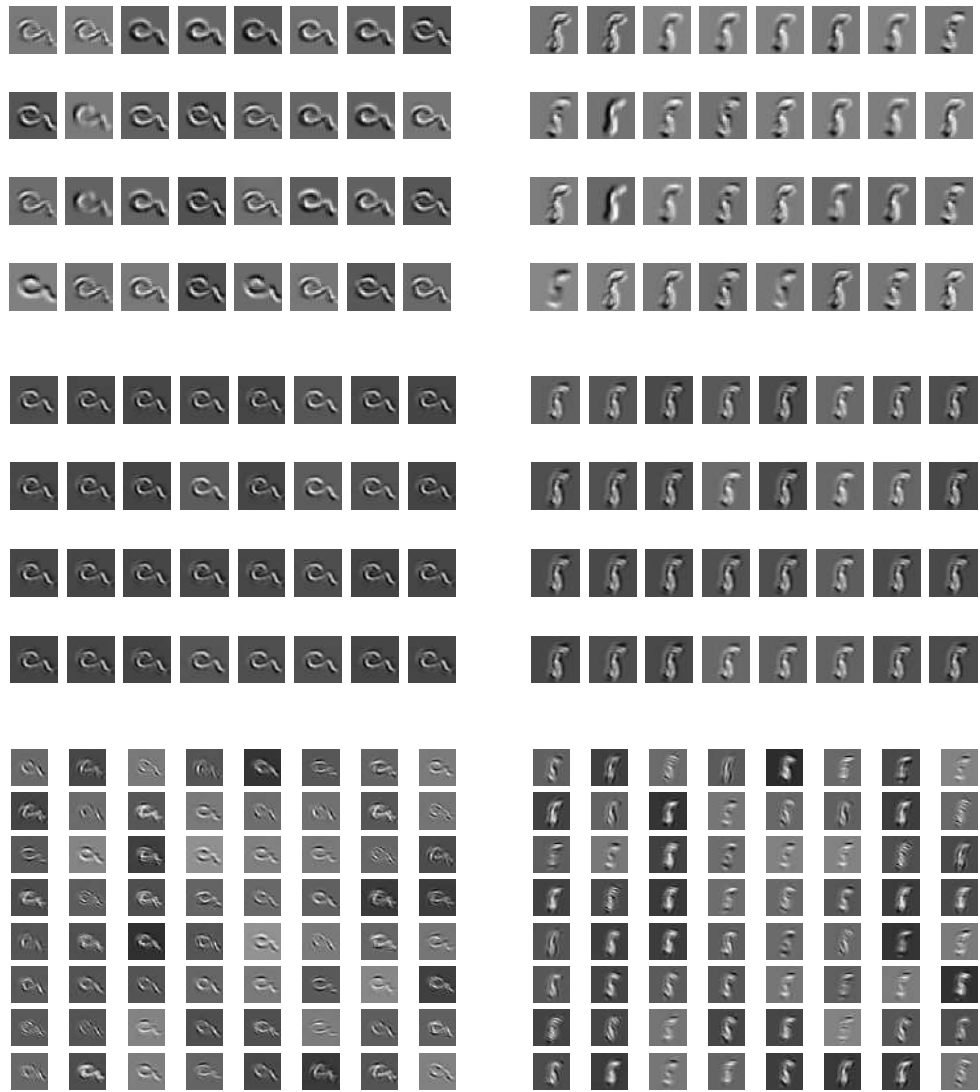

Figure 7: Example hierarchical features extracted at layers 1,2, and 3 for a GHN trained with MNIST classification.

APPENDIX C: DEEP EPITOMES WITH CIFAR10 IMAGE CLASSIFICATION

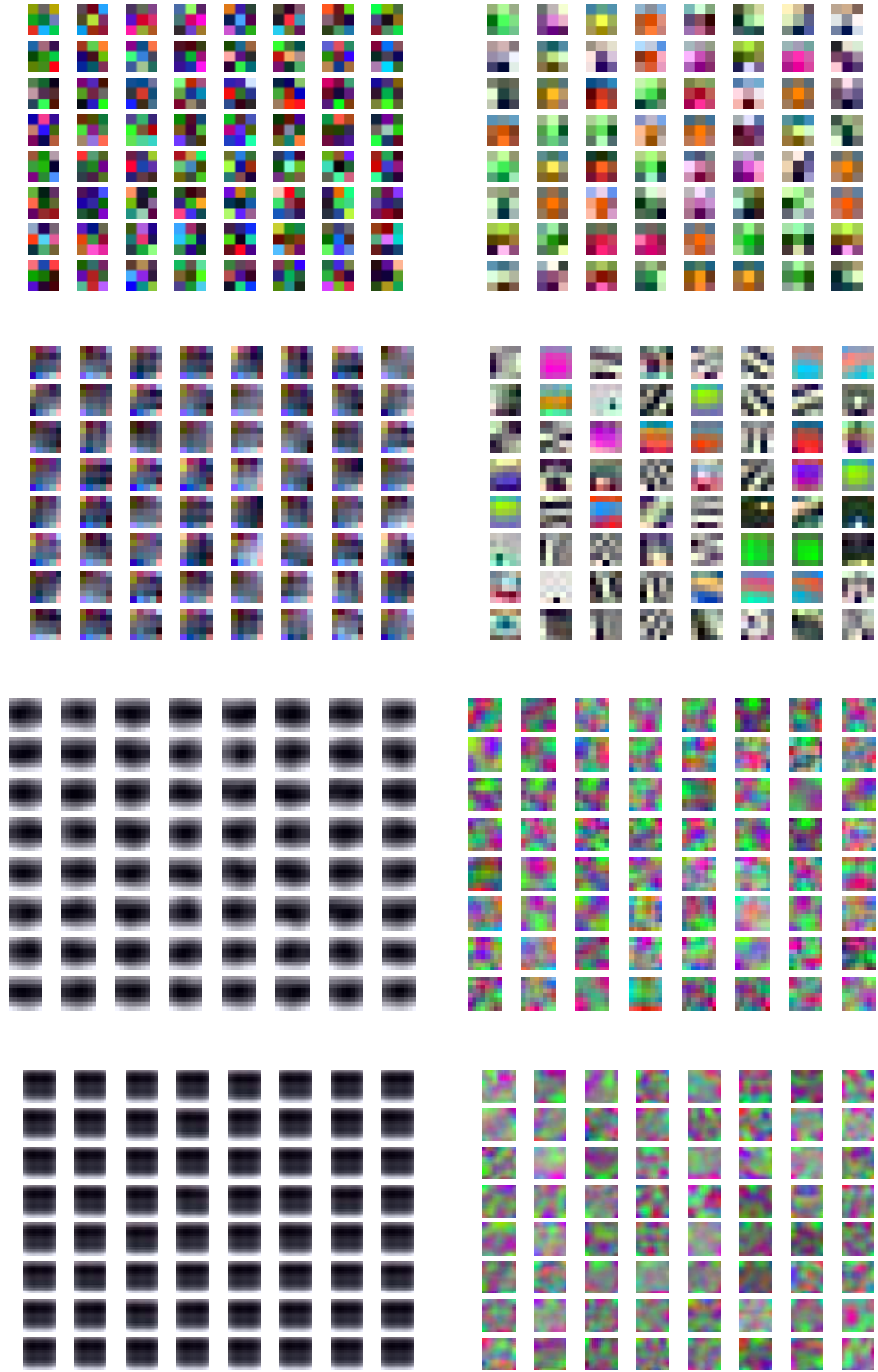

Figure 8: Top to bottom: deep epitomes for first 64 filters at layers 1,2, 3 and 4 of a GHN trained with CIFAR10 classification. Pseudo colour images correspond to three channels of merged epitomes from the first layer filters. **Left** column: iteration 100; **Right** column: iteration 180000.

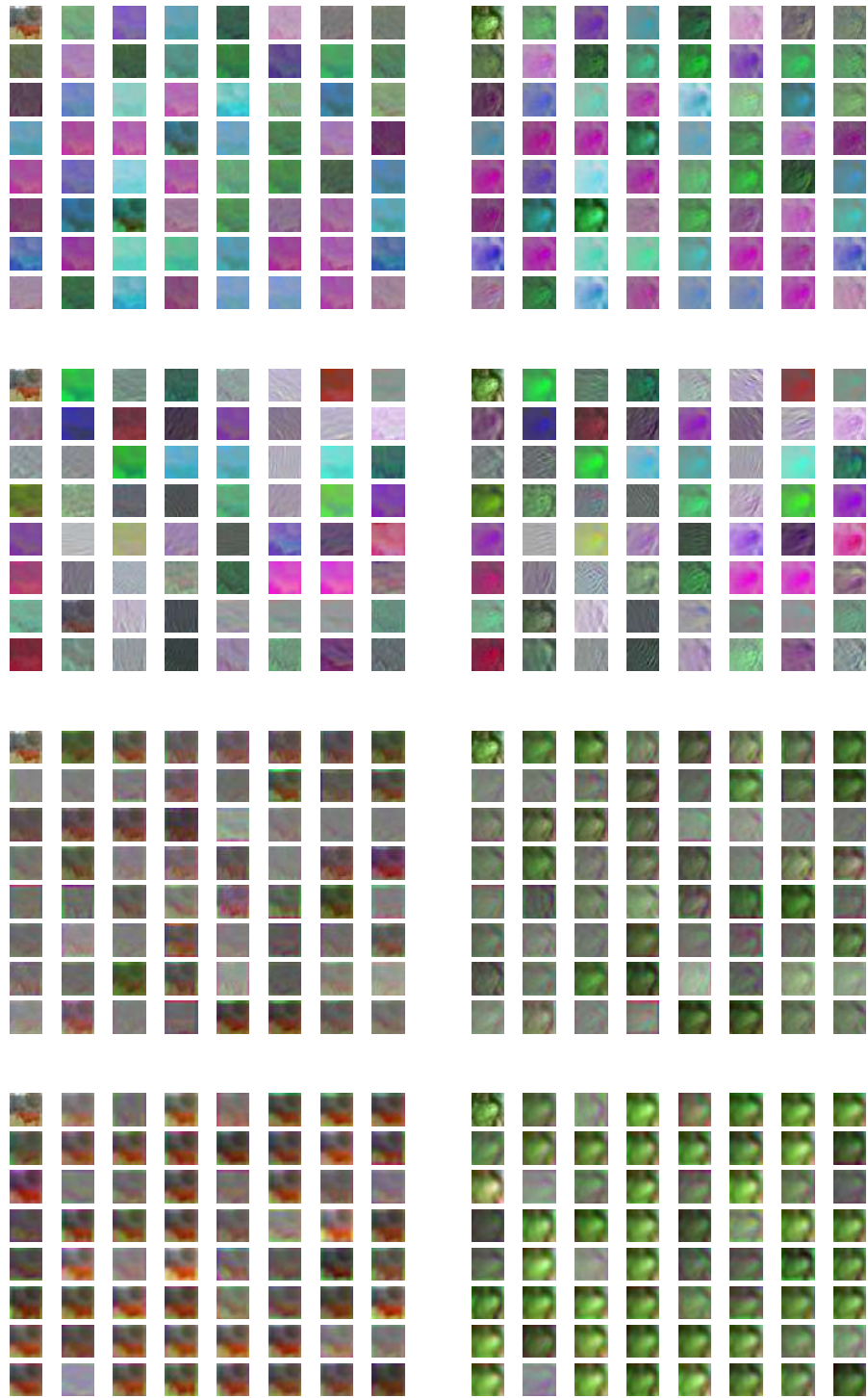

Figure 9: Hierarchical features extracted at layers 1,2,3 and 4 for a GHN trained with CIFAR10 at 180000 iterations. The top-left most image in each panel is the input image, and the rest are features extracted with different epitomes (only first 63 features are shown for layer 4). Pseudo colour images correspond to three channels of features outputs for input RGB colour channels. Note that oriented edgelets (layer 1,2), textons with associated colours (layer 2,3) and rough segmentations (layer 4) are extracted from different layers.

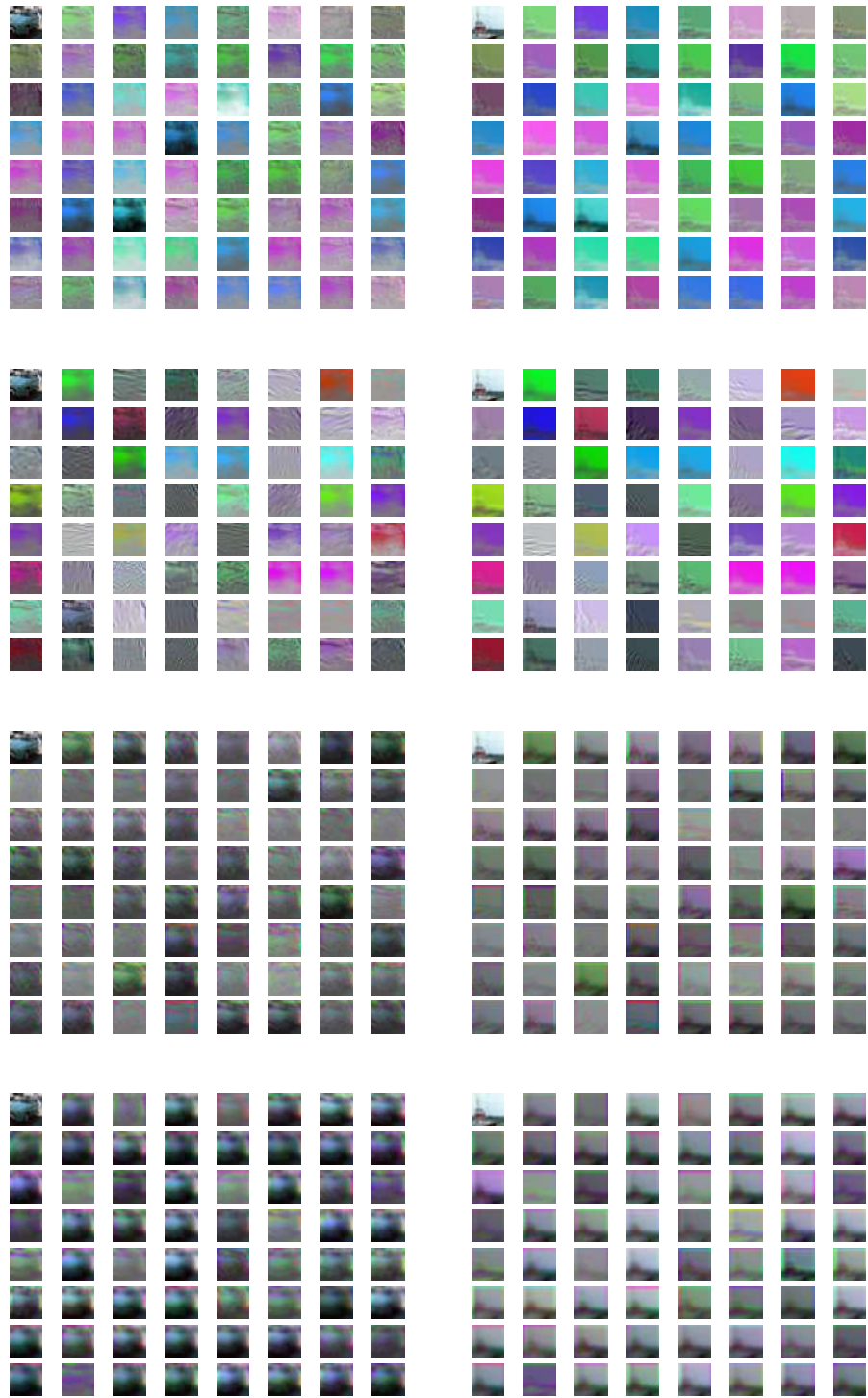

Figure 10: Hierarchical features extracted at layers 1,2,3 and 4 for a GHN trained with CIFAR10 at 180000 iterations. The top-left most image in each panel is the input image, and the rest are features extracted with different epitomes (only first 63 features are shown for layer 4). Pseudo colour images correspond to three channels of features outputs for input RGB colour channels. Note that oriented edgelets (layer 1,2), textons with associated colours (layer 2,3) and rough segmentations (layer 4) are extracted from different layers.

APPENDIX D: DEEP EPITOMES WITH CIFAR100 IMAGE CLASSIFICATION

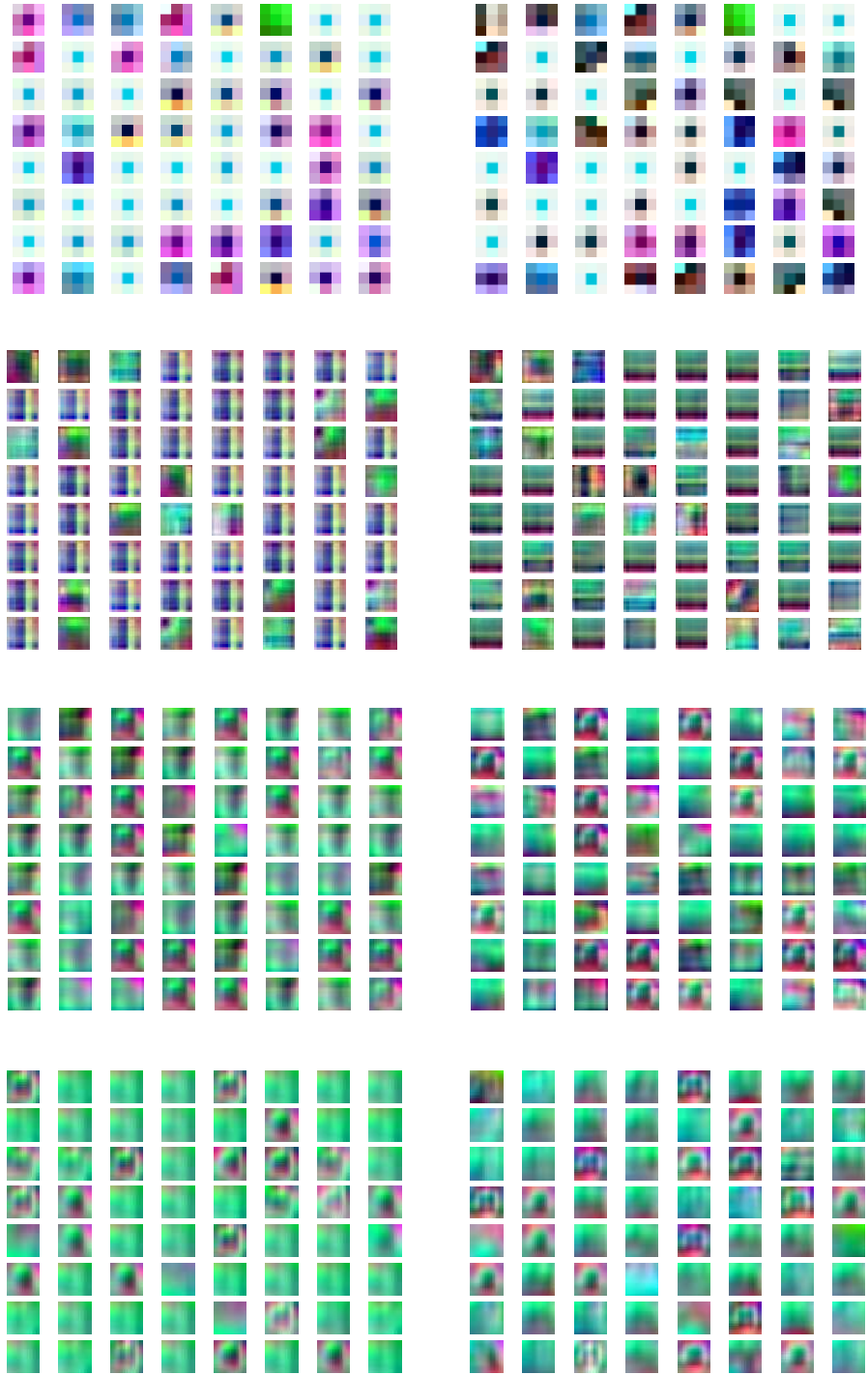

Figure 11: Top to bottom: deep epitomes for first 64 filters at layers 1,2, 3 and 4 of a GHN trained with CIFAR100 classification. Pseudo colour images correspond to three channels of merged epitomes from the first layer filters. **Left** column: iteration 10000; **Right** column: iteration 30000.

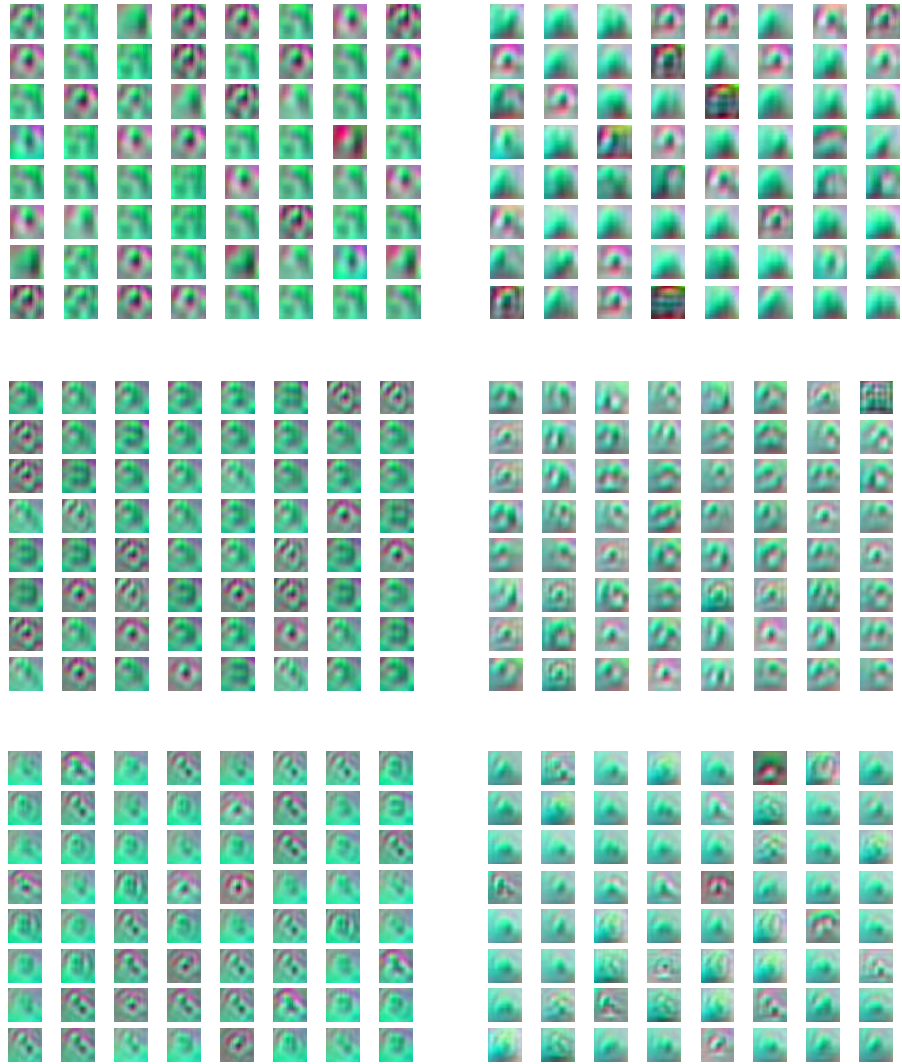

Figure 12: Top to bottom: deep epitomes for first 64 filters at layers 5, 6 and 7 of a GHN trained with CIFAR100 classification. Pseudo colour images correspond to three channels of merged epitomes from the first layer filters. **Left** column: iteration 10000; **Right** column: iteration 30000.

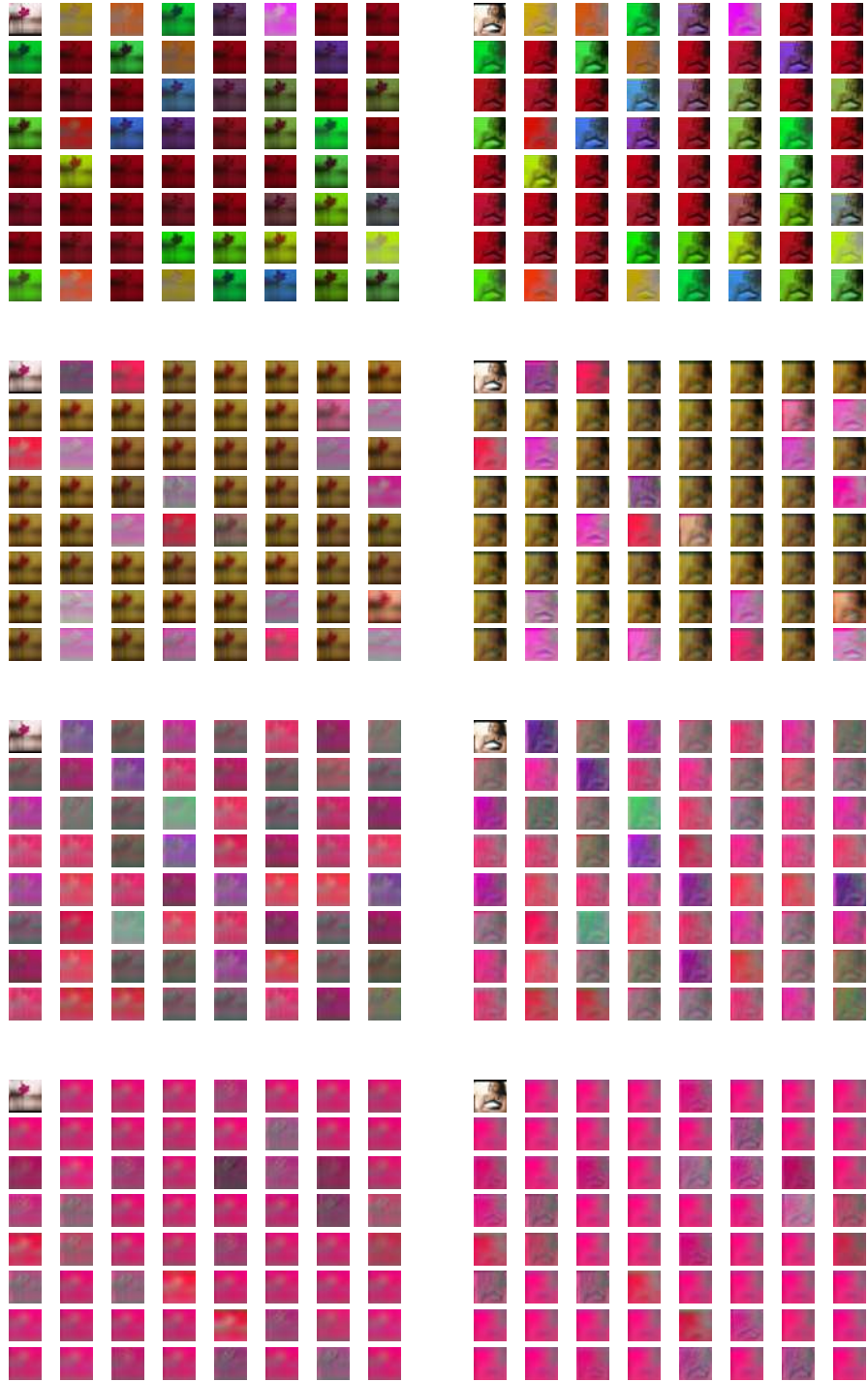

Figure 13: Hierarchical features extracted at layers 1,2,3 and 4 for a GHN trained with CIFAR100 at 10000 iterations. The top-left most image in each panel is the input image, and the rest are features extracted with different epitomes (only first 63 features are shown for different layers). Pseudo colour images correspond to three channels of features outputs for input RGB colour channels.

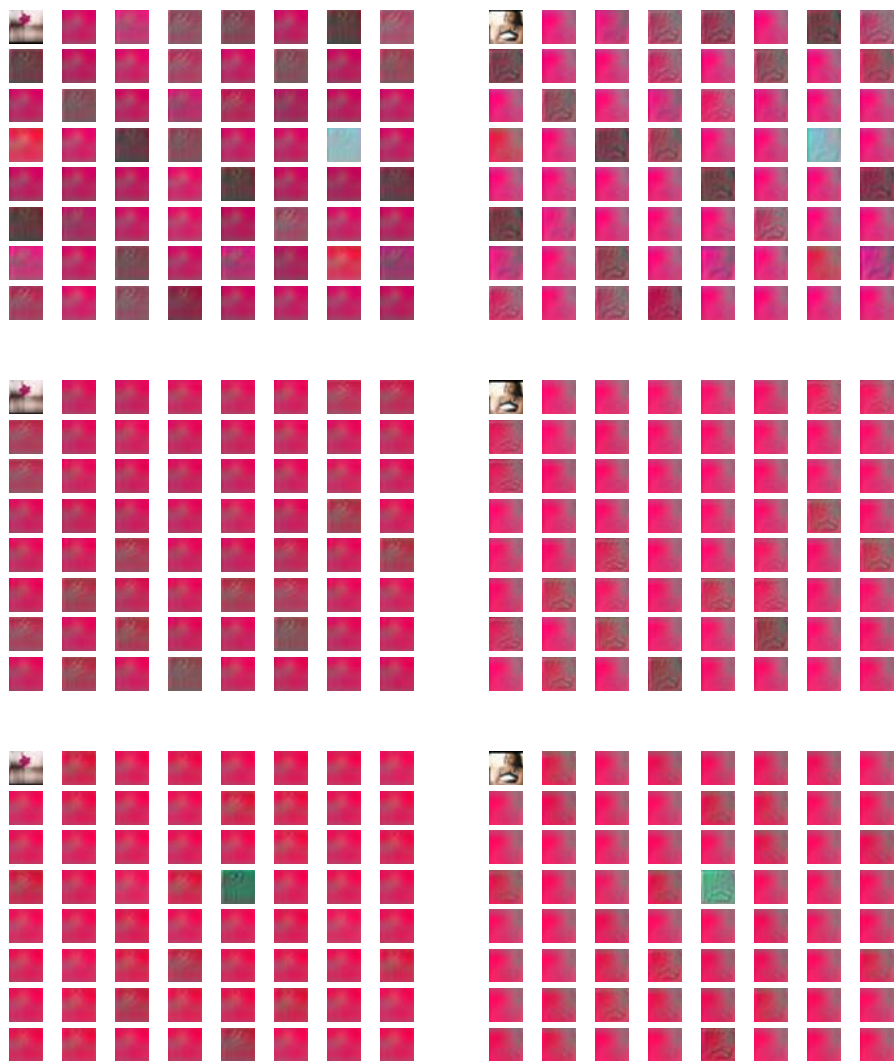

Figure 14: Hierarchical features extracted at layers 5,6 and 7 for a GHN trained with CIFAR100 at 10000 iterations. The top-left most image in each panel is the input image, and the rest are features extracted with different epitomes (only first 63 features are shown for different layers). Pseudo colour images correspond to three channels of features outputs for input RGB colour channels.